\documentclass[11pt]{article}

\usepackage{acl}

\definecolor{tblHead}{HTML}{EEECE6}  
\definecolor{tblSOTA}{HTML}{DCE3EA}  
\usepackage{caption}
\usepackage{subcaption}

\usepackage{times}
\usepackage{microtype}
\usepackage{xcolor} 
\usepackage[table]{xcolor}   
\usepackage{tabularx}        
\usepackage{booktabs}        
\definecolor{header}{HTML}{F2F2F2}
\definecolor{ours}{HTML}{DCE6F1}
\definecolor{best}{HTML}{D9EAD3}
\definecolor{second}{HTML}{FFF2CC}

\usepackage{amsmath}    
\usepackage{amssymb}    
\usepackage{amsfonts}  
\usepackage{subcaption}
\usepackage{graphicx}

\usepackage{algorithm}
\usepackage{algorithmic}

\usepackage{graphicx}
\usepackage{booktabs}
\usepackage{multirow}

\usepackage{enumitem}

\usepackage{xcolor}
\usepackage{caption}

\usepackage[T1]{fontenc}

\usepackage[utf8]{inputenc}

\usepackage{microtype}

\usepackage{inconsolata}

\usepackage{graphicx}

\title{Implicit Graph, Explicit Retrieval: Towards Efficient and Interpretable Long-horizon Memory for Large Language Models}

\author{
Xin Zhang$^{\dagger}$,
Kailai Yang$^{\dagger}$\thanks{~~Corresponding author.},
Hao Li$^{\ddagger}$,
Chenyue Li$^{\S}$,
Qiyu Wei$^{\dagger}$,
Sophia Ananiadou$^{\dagger}$ \\
$^{\dagger}$ University of Manchester \quad
$^{\ddagger}$ Imperial College London \quad
$^{\S}$ Stanford University \\
\texttt{\{xin.zhang-2, kailai.yang, qiyu.wei, sophia.ananiadou\}@manchester.ac.uk} \\
\texttt{lihao950220@hotmail.com} \quad
\texttt{chenyuel@stanford.edu}
}

\begin{document}
\maketitle

\begin{abstract}
Long-horizon applications increasingly require large language models (LLMs) to answer queries when relevant evidence is sparse and dispersed across very long contexts. Existing memory systems largely follow two paradigms: explicit structured memories offer interpretability but often become brittle under long-context overload, while latent memory mechanisms are efficient and stable yet difficult to inspect.
We propose \textsc{LatentGraphMem}, a memory framework that combines implicit graph memory with explicit subgraph retrieval. \textsc{LatentGraphMem} stores a graph-structured memory in latent space for stability and efficiency, and exposes a task-specific subgraph retrieval interface that returns a compact symbolic subgraph under a fixed budget for downstream reasoning and human inspection. During training, an explicit graph view is materialized to interface with a frozen reasoner for question-answering supervision. At inference time, retrieval is performed in latent space and only the retrieved subgraph is externalized.
Experiments on long-horizon benchmarks across multiple model scales show that \textsc{LatentGraphMem} consistently outperforms representative explicit-graph and latent-memory baselines, while enabling parameter-efficient adaptation and flexible scaling to larger reasoners without introducing large symbolic artifacts.

\end{abstract}

\section{Introduction}

\begin{figure*}[t]
  \centering
  \begin{subfigure}[t]{0.41\textwidth}
    \centering
    \includegraphics[width=\linewidth]{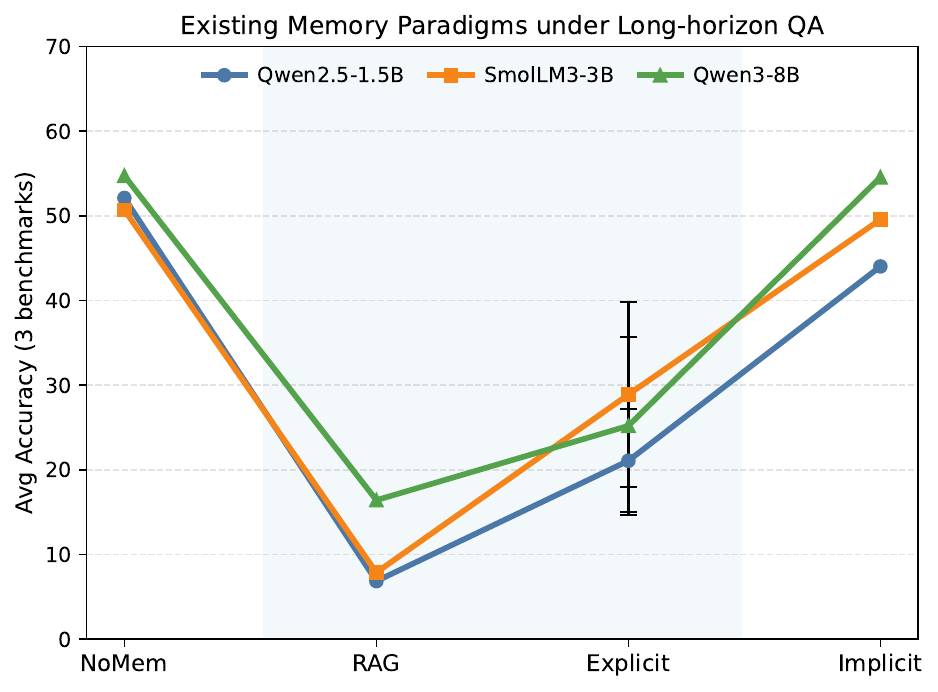}
    \subcaption{
    }
    \label{fig:intro_left}
  \end{subfigure}\hfill
  \begin{subfigure}[t]{0.53\textwidth}
    \centering
    \includegraphics[width=\linewidth]{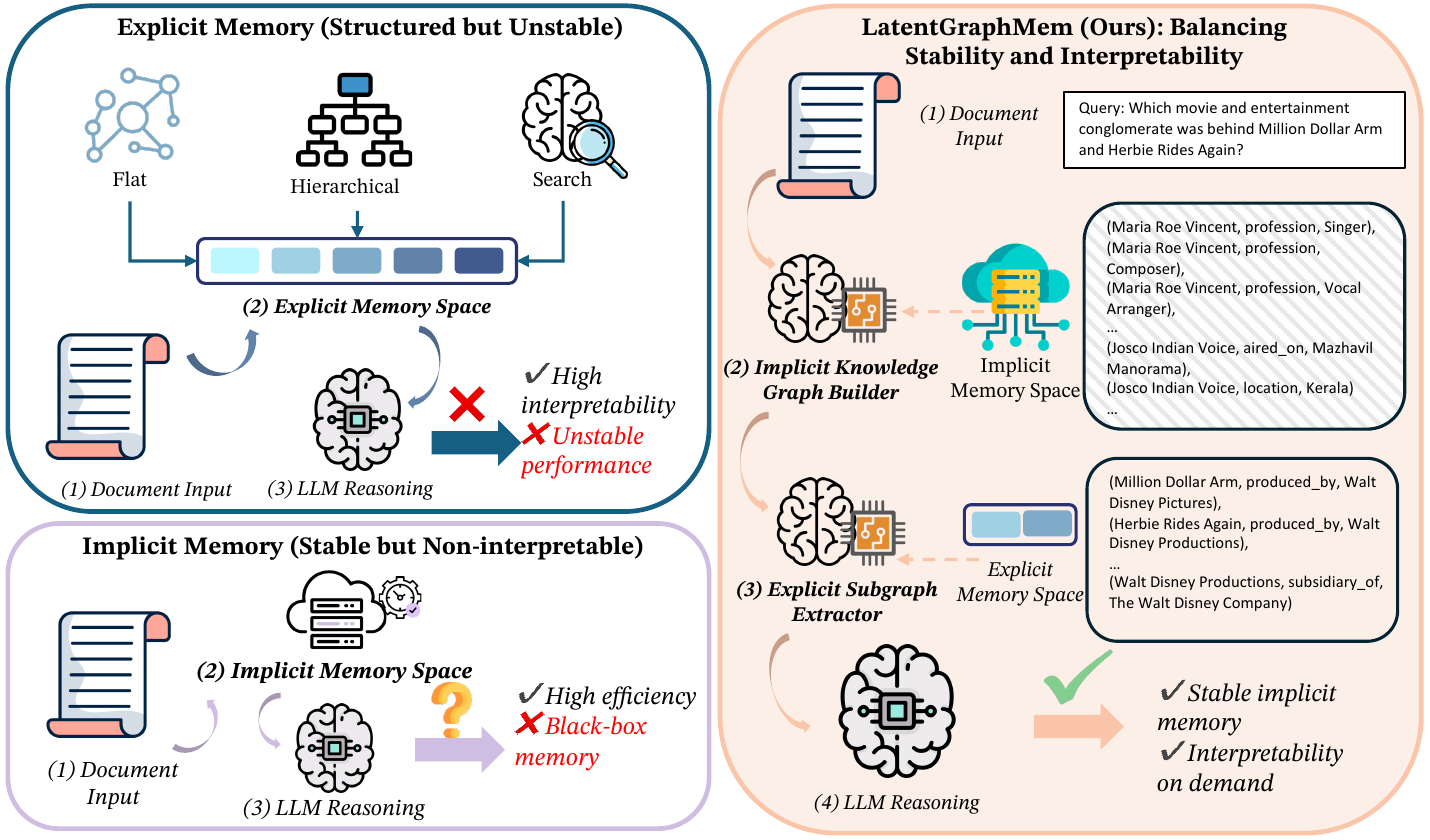}
      \subcaption{
    }
    \label{fig:intro_right}
  \end{subfigure}
  \caption{
  (a)     Performance of existing memory paradigms under long-horizon QA.
    \textbf{NoMem}: reasoner-only;
    \textbf{RAG}: retrieval-augmented generation;
    \textbf{Explicit}: explicit structured memory systems (A-Mem, PREMem, THEANINE, Mem0);
    \textbf{Implicit}: latent memory (MemGen). (b) Explicit and implicit memory paradigms for long-horizon reasoning.
  Existing explicit structured memory systems are interpretable but brittle under long and dispersed evidence,
  while implicit memory is more stable but opaque.
  \textsc{LatentGraphMem} combines implicit graph memory with explicit subgraph exposure.
  }
  \label{fig:intro_paradigm}
\end{figure*}

Memory mechanisms have become a foundational component of modern large language model (LLM) systems~\cite{hu2025memory}. In applications such as long-chain reasoning and multi-agent settings, carefully designed memory architectures enable LLMs to accumulate, organize, and reuse complex information beyond the constraints of a single prompt or fixed context window. Existing approaches can be broadly categorized into two dominant paradigms. The first paradigm consists of explicitly structured memory designs~\citep{DBLP:conf/euromlsys/NoohiDB25}, with particular emphasis on graph-structured representations~\citep{DBLP:journals/corr/abs-2511-01448,DBLP:journals/corr/abs-2504-19413,DBLP:conf/naacl/OngKGCKJHLY25}. The second paradigm encompasses implicit memory mechanisms, which store and manipulate information directly within continuous latent representations~\citep{DBLP:journals/corr/abs-2509-24704,wang2025mem,DBLP:conf/icml/0170KH0ZMGFH25}. Graph-based explicit memory mechanisms offer notable advantages in interpretability, as they facilitate inspection, debugging, and task-relevant memory graph retrieval~\citep{rasmussen2025zep,DBLP:conf/ijcai/AnokhinSSEK0B25,DBLP:conf/nips/GutierrezS0Y024}. In contrast, latent memory mechanisms demonstrate high efficiency, requiring substantially lower storage and computational overhead during memory usage. 

In this work, we focus on the important long-horizon memory scenarios\citep{DBLP:conf/acl/BaiTZ0WLCX0D0L25,DBLP:journals/corr/abs-2507-22411}, where LLMs must operate over multi-thousand-token contexts, in which task-relevant evidence is sparsely scattered across distant segments of the memory.
To compare the two memory paradigms, we conduct preliminary experiments on three representative long-horizon memory benchmarks and evaluate five representative memory mechanisms, including MemGen\citep{DBLP:journals/corr/abs-2509-24704}, A-Mem\citep{xu2025mem}, PREMem\citep{DBLP:journals/corr/abs-2509-10852}, THEANINE\citep{DBLP:conf/naacl/OngKGCKJHLY25}, and Mem0\citep{DBLP:journals/corr/abs-2504-19413}. As illustrated in Figure \ref{fig:intro_left}, explicit structured memory approaches frequently underperform even a vanilla baseline memory system across multiple tasks.  This instability is primarily caused by long-context overload: (i) structure induction becomes unreliable when the input is long and noisy, (ii) retrieval becomes increasingly sensitive to query variations as the symbolic space grows. In contrast, implicit memory mechanisms yield more consistent improvements. Nevertheless, their principal limitation lies in low interpretability, as it remains challenging to inspect the stored representations or to explain why specific pieces of evidence are retrieved during task execution.


The above analysis highlights the limitations of relying exclusively on either fully explicit structured memories or purely latent memory paradigms in long-horizon settings. Motivated by these observations, we revisit a practical assumption in memory design: while an explicit symbolic view is often needed to support interpretable evidence access for downstream reasoning, operating on explicit structured alone can be brittle and inefficient under long and noisy contexts.
We propose \textsc{LatentGraphMem}, a simple yet effective memory framework that combines the complementary strengths of both paradigms: it learns a graph memory with latent representations for efficient retrieval and control, while enabling task-specific \emph{explicit} subgraph exposure to support interpretability. Figure~\ref{fig:intro_right} illustrates the conceptual advantages of \textsc{LatentGraphMem} over previous approaches.
\textsc{LatentGraphMem} consists of two core components. First, a \textbf{graph memory builder} is trained via parameter-efficient adaptation to construct a bounded symbolic graph view from long documents in a streaming manner, and to produce latent embeddings for the resulting graph edges. Second, a \textbf{subgraph memory retriever} is adapted to select a compact evidence subgraph from these latent edge embeddings under a fixed retrieval budget, which is then externalized and fed to a frozen reasoner for answering. Notably, \textsc{LatentGraphMem} performs retrieval in latent space after the builder is trained, and at inference time it exposes only a compact, interpretable evidence subgraph to the reasoner rather than the full graph.

Extensive experiments on three long-horizon QA benchmarks demonstrate that \textsc{LatentGraphMem} consistently outperforms state-of-the-art graph-based and latent memory baselines across frozen-reasoner scales ranging from \emph{1.5B} to \emph{8B}. In particular, \textsc{LatentGraphMem} achieves the best average accuracy at every scale, reaching 56.08\%, 58.64\%, and 63.34\% under Qwen2.5-1.5B, SmolLM3-3B, and Qwen3-8B, respectively, surpassing the strongest competing memory system. These results confirm that encoding graph-structured memory implicitly while exposing compact explicit subgraphs enables robust and scalable long-horizon reasoning. Overall, this paper makes three primary contributions: (1) a systematic empirical study of explicit and implicit memory paradigms in long-horizon settings; (2) \textsc{LatentGraphMem}, a simple framework that unifies latent graph storage with explicit subgraph retrieval; and (3) consistent performance gains across model scales while maintaining interpretability through explicit retrieved evidence.

\section{Related Work}

\subsection{Explicit Structured Memory and Graph-Based Retrieval}
A dominant line of agent memory research builds an \emph{explicit} external store (often text, key--value entries, or databases) and retrieves from it at test time.
Representative systems include flat or database-style memories such as MemGPT~\citep{DBLP:journals/corr/abs-2310-08560}, as well as \emph{structured} memories that impose relational and temporal organization to support controlled retrieval and updates.
Recent graph-oriented agent memories maintain entity- or chunk-level nodes with explicit links, including production-focused long-term memory systems with an optional graph variant (Mem0 / Mem0g)~\citep{DBLP:journals/corr/abs-2504-19413} and dynamic indexing/linking designs (A-Mem)~\citep{xu2025mem} that build an explicit memory network for retrieval-time access~\citep{xu2025mem}.
Beyond general graph stores, several works emphasize \emph{graph-based retrieval} with higher-level structure: GraphRAG~\citep{DBLP:journals/corr/abs-2404-16130} constructs a multi-level graph index via community detection and recursive summarization, and Zep~\citep{rasmussen2025zep} models agent memory as a temporal knowledge graph with community partitioning.
For long-term dialogue, THEANINE~\citep{DBLP:conf/naacl/OngKGCKJHLY25} organizes stored experiences along explicit temporal links and retrieves coherent timelines rather than isolated top-$k$ items.
PREMem further strengthens explicit memory construction by performing pre-storage reasoning to form higher-quality structured memories for downstream retrieval and generation~\citep{DBLP:journals/corr/abs-2509-10852}.
These explicit-structure approaches are interpretable, but their end-to-end performance is highly sensitive to errors in extraction, indexing, and retrieval, and they require re-serializing structured artifacts into the token space for reasoning.

\subsection{Latent and Implicit Memory in the Representation Space}
In parallel, a second family of approaches stores memory implicitly in the model's internal representations rather than human-readable tokens.
A recent survey terms this latent memory, and organizes methods by how the latent state is formed and introduced, including Generate, Reuse, and Transform mechanisms~\citep{hu2025memory}.
Representative Generate methods compress long-horizons into learnable soft representations, such as AutoCompressor~\citep{DBLP:conf/emnlp/ChevalierWAC23} and MemoRAG~\citep{DBLP:conf/www/Qian0ZMLD025}, while MemoryLLM maintains persistent latent tokens for factual memory~\citep{DBLP:conf/icml/WangGCJLYYLLYSM24}.
More architecturally integrated designs extend latent memory across layers (e.g., M+) or introduce structured latent slots (e.g., LM2)~\citep{DBLP:conf/icml/0170KH0ZMGFH25,kang2025lm2}.
A different branch internalizes memory production into parameter dynamics, e.g., Titans~\citep{DBLP:journals/corr/abs-2501-00663}.
Closer to our setting, MemGen dynamically generates latent memory during decoding via lightweight adaptation modules~\citep{DBLP:journals/corr/abs-2509-24704}.
Compared with explicit stores, latent memory can preserve fine-grained information and avoid excessive prompt expansion, but it remains largely opaque and difficult to interpret.

\begin{figure*}[tp]
\centering
\includegraphics[scale=0.47]{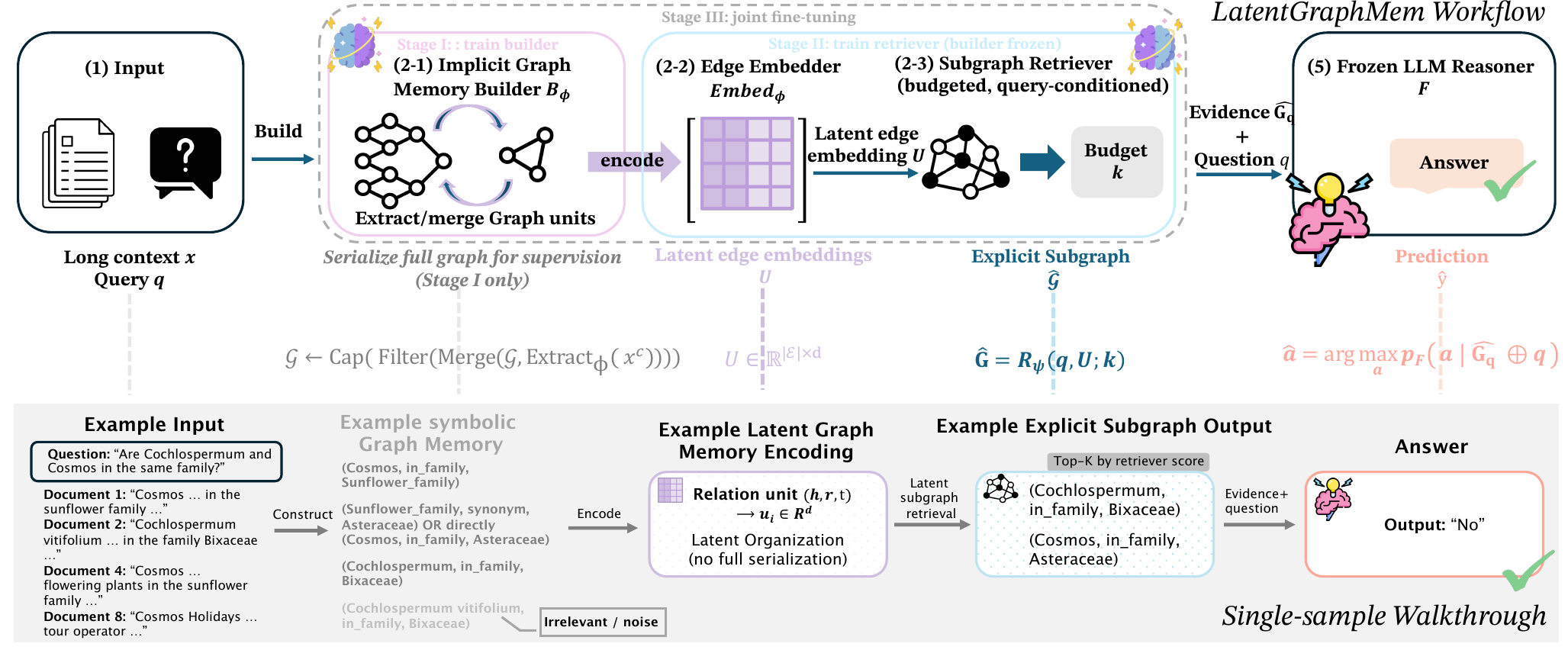}
\caption{
\textbf{Overview of \textsc{LatentGraphMem}.}
Given a long-horizon $x$ and query $q$, \textsc{LatentGraphMem} builds an implicit graph-structured memory from $x$, extracts a compact explicit subgraph as evidence conditioned on $q$, and answers with a frozen LLM reasoner using only the question and the extracted subgraph.
The context is used only for memory construction, and the lower panel shows a single-sample walkthrough.
}
\label{fig:lgm_overview}
\end{figure*}

\section{Method}

We propose \textsc{LatentGraphMem}, a three-stage framework for long-context question answering that leverages a learned graph memory to support reasoning, corresponding to graph construction, subgraph retrieval, and joint refinement. Given a long document \(x\) and a question \(q\), our goal is to construct and utilize structured evidence that enables a pretrained language model to generate accurate answers. \textsc{LatentGraphMem} consists of two trainable agents and a frozen reasoner. A graph builder constructs a global graph representation from the document, while a subgraph retriever selects question-relevant parts of this graph under a fixed budget. The selected evidence is then provided to an LLM for answer generation. 

Formally, the frozen reasoner \(F\) defines a conditional distribution \(p_F(\cdot \mid \cdot)\). We use \(\oplus\) to denote string concatenation (i.e. prompt composition). The builder \(B_\phi\) outputs an edge set \(\mathcal{E}\) (triples) from \(x\), and the retriever \(R_\psi\) selects a subgraph \(S_q\subseteq \mathcal{E}\) conditioned on \(q\). Each training instance is $(x,q,a)$, where \(a = (a_1,\ldots,a_{|a|})\) denotes the gold answer token sequence.

\subsection{Stage I: Remote-Supervised Full-Graph Construction }

\paragraph{Streaming graph construction for long documents.}
Context in long-horizon scenario \(x\) is usually long, so we split it into overlapping chunks \(\{x^{(c)}\}_{c=1}^{C}\) with maximum length \(L\) tokens (overlap \(O\)). The builder processes chunks sequentially and maintains an evolving explicit graph state:
\begin{equation}
\begin{aligned}
\mathcal{G}^{(0)} &= \emptyset, \\
Y^{(c)} &= \mathrm{Extract}_{\phi}(x^{(c)}),\\
\tilde{\mathcal{G}}^{(c)} &= \mathrm{Merge}\!\left(\mathcal{G}^{(c-1)},\, Y^{(c)}\right),\\
\mathcal{G}^{(c)} &= \mathrm{Cap}\!\left(\mathrm{Filter}\!\left(\tilde{\mathcal{G}}^{(c)}\right)\right).
\end{aligned}
\end{equation}
\(\mathrm{Extract}_{\phi}\) emits a small set of candidate triples from chunk \(x^{(c)}\). 
We denote the edge set of $\mathcal{G}^{(c)}$ as $\mathcal{E}^{(c)}$. 
After processing all chunks, we define the final graph as $\mathcal{G}=\mathcal{G}^{(C)}$ and its edge set as $\mathcal{E}=\mathcal{E}^{(C)}$.
\(\mathrm{Merge}\) canonicalizes and deduplicates entities/relations across chunks. \(\mathrm{Filter}\) enforces quality constraints (e.g., schema validity, field-length limits, duplicate removal, and relation-type constraints). \(\mathrm{Cap}\) enforces a hard capacity limit to prevent graph explosion. Concretely, $\mathrm{Cap}$ enforces $|\mathcal{E}^{(c)}|\le M$ at every step $c$.
For each chunk, the builder emits a constrained triple set:
\begin{equation}
Y^{(c)}=\{(h_j,r_j,t_j)\}_{j=1}^{|Y^{(c)}|},~ |Y^{(c)}|\le m_{\text{chunk}},
\end{equation}
where \(m_{\text{chunk}}\) is a fixed per-chunk extraction cap. We additionally apply hard constraints: (i) maximum token length for each field \(h,r,t\), and (ii) a global edge cap \(M\) such that the final explicit graph \(\mathcal{G}=\mathcal{G}^{(C)}=(\mathcal{V},\mathcal{E})\) satisfies \(|\mathcal{E}|\le M\). Here $\mathcal{V}$ denotes the set of unique entities appearing in $\mathcal{E}$. These controls ensure stable graph size and maintain a consistent evidence budget.

\paragraph{Reasoner interface (explicit full graph) and QA-only SFT loss.}
Stage~I trains the builder using only the downstream answer objective. We serialize the explicit full graph into a deterministic evidence string:
$\hat{G}=\mathrm{Serialize}(\mathcal{E}),$
where \(\mathrm{Serialize}\) formats edges using a fixed template (e.g., \texttt{Relevant Knowledge: [h|r|t] ...}). 
Serializing the full graph \(\hat{G}\) is used only in Stage~I to interface with the reasoner for supervision; at inference time, we serialize only the retrieved subgraph \(S_q\).
The frozen reasoner \(F\) defines an autoregressive conditional distribution \(p_F(\cdot)\) over answer tokens conditioned on the serialized graph evidence and the question.
We optimize the builder parameters \(\phi\) by minimizing the cross-entropy loss:
\begin{equation}
\mathcal{L}_{\mathrm{I}}(\phi)= -\sum_{t=1}^{|a|}\log p_{F}\!\left(a_t \mid a_{<t},\, \hat{G}\oplus q\right),
\end{equation}
where \(a_{<t}=(a_1,\ldots,a_{t-1})\) denotes the prefix of the gold answer under teacher forcing.

\subsection{Stage II: Remote-Supervised Latent Subgraph Retrieval }

\paragraph{Implicit graph representation after Stage I.}
Although Stage~I materializes an explicit graph to interface with the reasoner, after training we use an implicit representation for downstream control. Specifically, each retained edge \(e_i=(h_i,r_i,t_i)\in\mathcal{E}\) is mapped to a latent embedding:
\begin{equation}
u_i=\mathrm{Embed}_{\phi}(h_i,r_i,t_i)\in\mathbb{R}^{d},~i=1,\ldots,|\mathcal{E}|.
\end{equation}
We store the edge embeddings as a matrix $U\in\mathbb{R}^{|\mathcal{E}|\times d}$, 
where the $i$-th row corresponds to edge $e_i$ and equals $u_i$. Stage~II and Stage~III perform selection in this latent space.

\paragraph{Training regime and inputs/outputs.}
Stage~II starts from the builder trained in Stage~I and fixes \(B_\phi\). Given \((q, U)\), the subgraph retriever \(R_\psi\) selects a compact subset of edges under a budget \(k\), operating only on the embeddings \(U\). The output is an index set \(\mathcal{I}_q\) and the corresponding edge subset \(S_q=\{e_i\mid i\in\mathcal{I}_q\}\).

\paragraph{Latent scoring and budgeted selection.}
We encode the question and compute relevance scores for each edge embedding:
\begin{equation}
v=R_{\psi}^{\mathrm{enc}}(q)\in\mathbb{R}^{d},~
s_i=v^{\top}W u_i,~ W\in\mathbb{R}^{d\times d}.
\end{equation}
The bilinear weight $W$ is a learnable parameter of the retriever and is included in $\psi$.
We then select at most \(k\) edges:
\begin{equation}
\mathcal{I}_q=\mathrm{TopK}(\{s_i\},k),~
S_q=\{e_i\mid i\in\mathcal{I}_q\}.
\end{equation}
This provides explicit budget control and yields a compact subgraph.

\paragraph{Differentiable training path (SFT, no RL).}
Since \(\mathrm{TopK}\) is not differentiable, we use a straight-through estimator: the forward pass uses the hard top-\(k\) selection, while the backward pass routes gradients through a softmax relaxation with temperature \(\tau\):
\begin{equation}
z=\mathrm{TopKMask}(\{s_i\},k),~
\alpha_i=\frac{\exp(s_i/\tau)}{\sum_{j}\exp(s_j/\tau)}.
\end{equation}
We form \(S_q\) from \(z\) in the forward pass and use \(\alpha\) for gradient flow in the backward pass. This keeps training within standard backpropagation and does not require reinforcement learning.

\definecolor{headGray}{HTML}{F4F4F4}
\definecolor{oursGray}{HTML}{EEEEEE}
\definecolor{bandRed}{HTML}{FAECEC}
\definecolor{bandOrg}{HTML}{FDF3EC}
\definecolor{bandBlu}{HTML}{EAF6FC}

\newcommand{\best}[1]{\textbf{#1}}
\newcommand{\second}[1]{\underline{#1}}
\newcommand{\worst}[1]{\textcolor{black!45}{#1}}

\begin{table*}[t]
\centering
\small
\setlength{\tabcolsep}{3.2pt}
\renewcommand{\arraystretch}{1.16}

\caption{Main results on the merged QA test set (Acc/ROUGE-L, higher is better; shown in \%). Avg is the unweighted mean over the three datasets. We highlight best/second/worst for each column (bold/underline/gray).}
\label{tab:merged_main_acc_rougel_final}

\begin{tabular}{>{\raggedright\arraybackslash}p{0.22\textwidth}
                cc !{\vrule width 0.4pt}
                cc !{\vrule width 0.4pt}
                cc !{\vrule width 0.4pt}
                cc}
\toprule
\rowcolor{headGray}
\multirow{2}{*}{\cellcolor{headGray}\textbf{Method}} &
\multicolumn{2}{c!{\vrule width 0.4pt}}{\textbf{HotpotQA}} &
\multicolumn{2}{c!{\vrule width 0.4pt}}{\textbf{NarrativeQA}} &
\multicolumn{2}{c!{\vrule width 0.4pt}}{\textbf{WikiHop}} &
\multicolumn{2}{c}{\textbf{Avg}} \\
\cmidrule(lr){2-3}\cmidrule(lr){4-5}\cmidrule(lr){6-7}\cmidrule(lr){8-9}
\rowcolor{headGray}
& \textbf{Acc} & \textbf{R-L}
& \textbf{Acc} & \textbf{R-L}
& \textbf{Acc} & \textbf{R-L}
& \textbf{Acc} & \textbf{R-L} \\
\midrule

\multicolumn{9}{c}{\cellcolor{bandRed}\textbf{Backbone: Qwen2.5-1.5B-Instruct}} \\
\specialrule{0.5pt}{2pt}{4pt}

Reasoner-only &
\second{80.20} & 52.20 &
41.50 & 28.13 &
\second{34.63} & 23.57 &
\second{52.11} & 34.63 \\

MemGen &
57.10 & \second{57.59} &
\second{42.63} & \best{54.20} &
32.37 & \best{32.54} &
44.03 & \best{48.11} \\

A-Mem &
46.60 & 56.73 &
15.62 & \second{49.85} &
12.12 & 20.53 &
24.78 & \second{42.37} \\

PREMem &
49.70 & 48.78 &
10.37 & 11.64 &
10.13 & 10.78 &
23.40 & 23.73 \\

THEANINE &
19.40 & 8.19 &
2.25 & 3.84 &
10.25 & 6.64 &
10.63 & 6.22 \\

Mem0 &
63.80 & 44.08 &
\worst{1.12} & \worst{5.23} &
11.63 & 10.15 &
25.52 & \worst{19.82} \\

RAG &
\worst{14.60} & \worst{29.05} &
4.50 & 26.47 &
\worst{1.50} & \worst{8.05} &
\worst{6.87} & 21.19 \\

\rowcolor{oursGray}
\textsc{LatentGraphMem (Ours)} &
\best{86.60} & \best{58.54} &
\best{45.50} & 32.25 &
\best{36.13} & \second{25.03} &
\best{56.08} & 38.61 \\

\specialrule{0.9pt}{4pt}{8pt}

\multicolumn{9}{c}{\cellcolor{bandOrg}\textbf{Backbone: SmolLM3-3B}} \\
\specialrule{0.5pt}{2pt}{4pt}

Reasoner-only &
\second{77.10} & 51.46 &
\second{40.50} & 37.43 &
\second{34.50} & 24.61 &
50.70 & 37.83 \\

MemGen &
58.00 & 53.64 &
49.38 & 47.37 &
41.35 & \best{42.59} &
\second{49.58} & \second{47.87} \\

A-Mem &
51.30 & \second{61.18} &
19.88 & \best{55.99} &
17.13 & 28.45 &
29.44 & \best{48.54} \\

PREMem &
87.70 & 60.00 &
11.22 & 9.11 &
38.50 & 25.77 &
45.81 & 31.63 \\

THEANINE &
42.79 & 19.93 &
10.87 & 7.84 &
19.75 & 13.36 &
24.47 & 13.71 \\

Mem0 &
44.90 & 29.30 &
\worst{0.25} & \worst{3.04} &
2.50 & \worst{2.26} &
15.88 & \worst{11.53} \\

RAG &
\worst{15.30} & 52.22 &
6.12 & \second{40.12} &
\worst{2.25} & 16.52 &
7.89 & 36.29 \\

\rowcolor{oursGray}
\textsc{LatentGraphMem (Ours)} &
\best{88.90} & \best{61.39} &
\best{47.00} & 35.38 &
\best{40.03} & \second{35.71} &
\best{58.64} & 44.16 \\

\specialrule{0.9pt}{4pt}{8pt}

\multicolumn{9}{c}{\cellcolor{bandBlu}\textbf{Backbone: Qwen3-8B}} \\
\specialrule{0.5pt}{2pt}{4pt}

Reasoner-only &
72.60 & 62.48 &
\best{51.00} & \second{63.81} &
\second{40.62} & 31.55 &
\second{54.74} & \second{52.61} \\

MemGen &
72.30 & \second{71.03} &
\second{50.88} & \best{63.94} &
40.50 & \second{40.30} &
54.56 & \best{58.42} \\

A-Mem &
31.40 & 55.10 &
15.88 & 46.04 &
9.38 & 17.89 &
18.89 & 39.68 \\

PREMem &
\second{81.46} & 56.53 &
2.57 & \worst{3.00} &
45.87 & 31.55 &
43.30 & 30.36 \\

THEANINE&
39.94 & 16.02 &
6.63 & 5.13 &
16.15 & 10.48 &
20.91 & 10.54 \\

Mem0 &
48.60 & \worst{29.43} &
\worst{0.50} & 3.12 &
\worst{4.13} & \worst{3.46} &
17.74 & \worst{12.00} \\

RAG &
\worst{29.30} & 63.88 &
14.87 & 47.29 &
5.12 & 20.04 &
16.43 & 43.74 \\

\rowcolor{oursGray}
\textsc{LatentGraphMem (Ours)} &
\best{90.20} & \best{71.21} &
\best{51.00} & 37.60 &
\best{48.81} & \best{40.37} &
\best{63.34} & 49.73 \\

\bottomrule

\end{tabular}
\label{tab:main_results}
\end{table*}

\paragraph{Reasoner interface (explicit retrieved subgraph) and QA-only SFT loss.}
We externalize only the retrieved subgraph for the reasoner:
\begin{equation}
\hat{G}_q=\mathrm{Serialize}(S_q),
\end{equation}
and train the retriever using the same QA-only cross-entropy loss:
\begin{equation}
\mathcal{L}_{\mathrm{II}}(\psi)= -\sum_{t=1}^{|a|}\log p_{F}\!\left(a_t \mid a_{<t},\, \hat{G}_q\oplus q\right).
\end{equation}

\subsection{Stage III: Joint Fine-Tuning }

\paragraph{Training regime.}
Stage~III starts from the builder obtained in Stage~I and the retriever obtained in Stage~II. We jointly fine-tune \((\phi,\psi)\) to maximize their coordination: the builder should produce edges/embeddings that are easy to retrieve and useful when externalized, and the retriever should best exploit the builder's latent space. The reasoner \(F\) remains frozen.

\paragraph{End-to-end dataflow and objective.}
The full computation is:
\begin{equation}
x \xrightarrow{B_{\phi}} (\mathcal{E},U) \xrightarrow{R_{\psi}(q)} S_q \xrightarrow{\mathrm{Serialize}} \hat{G}_q \xrightarrow{F} a,
\end{equation}
and we optimize the QA-only SFT objective:
\begin{equation}
\mathcal{L}_{\mathrm{III}}(\phi,\psi)= -\sum_{t=1}^{|a|}\log p_{F}\!\left(a_t \mid a_{<t},\, \hat{G}_q\oplus q\right).
\end{equation}
In practice, we use stable alternating updates: builder-only steps minimizing \(\mathcal{L}_{\mathrm{I}}\) to preserve Stage~I construction quality, and joint steps minimizing \(\mathcal{L}_{\mathrm{III}}\) to improve coupling.
Concretely, builder-only steps use the full-graph interface $\hat{G}=\mathrm{Serialize}(\mathcal{E})$ as an auxiliary objective to preserve extraction quality,
whereas joint steps use the retrieved subgraph interface $\hat{G}_q=\mathrm{Serialize}(S_q)$ to improve builder--retriever coupling.

\paragraph{Inference.}
At test time, given $(x,q)$, the builder streams over chunks and maintains an internal graph state, producing an edge set $\mathcal{E}$ and an embedding matrix $U\in\mathbb{R}^{|\mathcal{E}|\times d}$ without serializing the full graph into the prompt. Only the retrieved subgraph $S_q$ is serialized and passed to the frozen reasoner.
The retriever selects a budgeted subgraph \(S_q \subseteq \mathcal{E}\) in latent space, which is then serialized as evidence for the frozen reasoner:
\begin{equation}
\hat{a}=\arg\max_a p_F\!\left(a \mid \mathrm{Serialize}(S_q)\oplus q\right).
\end{equation}

\section{Experiments}
\label{sec:experiments}

\subsection{Datasets}
\label{sec:datasets}

We construct a long-horizon QA training corpus of 20{,}800 instances from three real-world sources (TriviaQA~\citep{joshi2017triviaqa}, QASPER~\citep{dasigi2021qasper}, and QuALITY~\citep{pang2022quality}) and evaluate on a fixed 2{,}600-instance mixture from HotpotQA~\citep{yang2018hotpotqa}, NarrativeQA~\citep{kocisky2018narrativeqa}, and WikiHop~\citep{welbl2018wikihop}.
Table~\ref{tab:datasets} reports dataset statistics computed on our processed splits, including character counts and tokenizer-specific token counts under each base model for consistent measurement across configurations. Additional statistics and dataset characteristics are provided in Appendix~\ref{sec:dataset_stats}.
\textbf{Training datasets.}
TriviaQA provides open-domain questions with long, heterogeneous evidence collections, QASPER is grounded in full scientific papers with structured long contexts, and QuALITY offers multiple-choice supervision over long articles, together covering diverse long-horizon reasoning patterns.
\textbf{Evaluation datasets.}
We evaluate only on HotpotQA, NarrativeQA, and WikiHop, where HotpotQA and WikiHop stress multi-hop reasoning across documents and entities and NarrativeQA probes long-form narrative understanding, and none of these evaluation datasets are included in training.

\subsection{Baselines}
\label{sec:baselines}
We compare \textsc{LatentGraphMem} with seven baselines. Reasoner-only (full context) answers directly from the full input (up to the context limit) without any external memory, while RAG retrieves top-$k$ chunks from the input via embedding similarity and appends them as textual evidence to the same frozen reasoner. In addition, we include five representative memory-system baselines covering both explicit structure and latent memory: THEANINE (timeline-style graph memory management) \citep{DBLP:conf/naacl/OngKGCKJHLY25}, PREMem (pre-storage reasoning for episodic memory) \citep{DBLP:journals/corr/abs-2509-10852}, Mem0 / Mem0g (production-oriented long-term memory with an optional graph variant) \citep{DBLP:journals/corr/abs-2504-19413}, A-Mem (agentic memory with dynamic indexing and linking) \citep{xu2025mem}, and MemGen (generative latent memory tokens) \citep{DBLP:journals/corr/abs-2509-24704}.

\paragraph{Implementation Details.}
For all experiments, the graph builder and subgraph retriever are instantiated as decoder-only LLMs with parameter-efficient LoRA adaptation, while the reasoner is kept frozen throughout training and inference. In Stage~I, long documents are processed in overlapping chunks of 1{,}024 tokens with an overlap of 128 tokens. We cap extraction to 32 triples per chunk and limit the global graph capacity to \(M=150\) edges to control memory growth. In Stage~II, the retriever selects a compact evidence subgraph under a fixed budget of \(k=30\) edges, which is serialized and provided to the frozen reasoner for answer generation. 
We report answer accuracy (Acc) and ROUGE-L on all benchmarks. 
The graph builder and subgraph retriever are instantiated with Qwen2.5-1.5B-Instruct, and we evaluate with frozen reasoners at 1.5B, 3B, and 8B scales by directly reusing the same trained memory modules without additional training.

\begin{figure}[t] 
  \centering
  \includegraphics[width=0.45\textwidth]{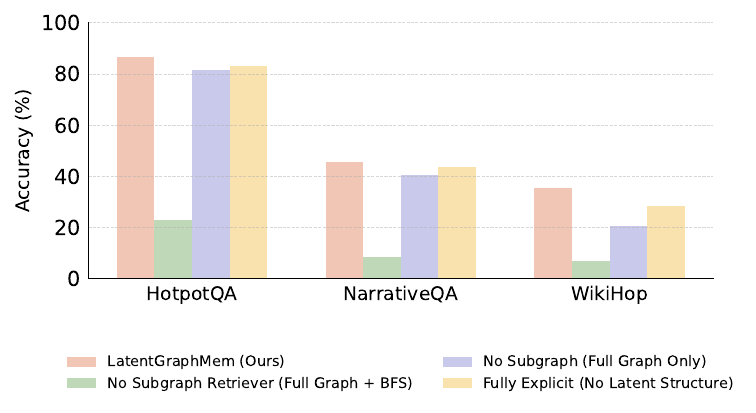}
  \caption{Memory Structure Ablation}
  \label{fig:graphstructure}
\end{figure}

\subsection{Main Results}
\label{sec:main_results}
We evaluate \textsc{LatentGraphMem} against seven baselines on three long-horizon QA benchmarks (HotpotQA, NarrativeQA, and WikiHop) under three frozen-reasoner backbones (Qwen2.5-1.5B, SmolLM3-3B, and Qwen3-8B), with results summarized in Table~\ref{tab:main_results}.
Across all settings, \textsc{LatentGraphMem} consistently achieves the strongest overall performance in terms of average accuracy.

Specifically, (1) \textsc{LatentGraphMem} improves average accuracy over all baselines at every model scale, raising the average accuracy to 56.08\% with Qwen2.5-1.5B and further to 63.34\% with Qwen3-8B, demonstrating that the proposed memory mechanism scales reliably with increasing reasoning capacity.
(2) Fully explicit graph-based memory methods, including THEANINE, PREMem, and Mem0/Mem0g, exhibit severe performance degradation in long-horizon settings, particularly on NarrativeQA, where they often collapse to near-zero or single-digit accuracy, resulting in substantially lower averages than both the reasoner-only baseline and latent-memory approaches.
(3) Compared to purely latent memory, represented by MemGen, \textsc{LatentGraphMem} achieves consistently higher accuracy on multi-hop benchmarks such as HotpotQA across all backbones, despite MemGen occasionally attaining higher ROUGE-L scores on NarrativeQA.
Notably, \textsc{LatentGraphMem} employs a single 1.5B graph builder that generalizes directly to larger frozen reasoners, whereas MemGen requires the memory module and reasoner to be trained at comparable scales to maintain latent alignment.



\definecolor{headGray}{HTML}{F4F4F4}
\definecolor{fastGray}{HTML}{EDEDED}

\begin{table}[!t]
\centering
\small
\setlength{\tabcolsep}{4pt}      
\renewcommand{\arraystretch}{1.05} 
\caption{Inference time (seconds) versus context length (k tokens) for long-context samples ($\ge$6k).}
\label{tab:time_vs_context}
\begin{tabular}{c c c c}
\toprule
\rowcolor{headGray}
Context (k) & Ours (s) & MemGen (s) & A-Mem (s) \\
\midrule
6k  & 12.47 & \cellcolor{fastGray}10.59 & 20.00 \\
7k  & 13.42 & \cellcolor{fastGray}11.15 & 17.48 \\
8k  & \cellcolor{fastGray}12.84 & 15.52 & 16.50 \\
9k  & 13.43 & \cellcolor{fastGray}10.86 & 44.65 \\
10k & 13.78 & \cellcolor{fastGray}8.95  & 41.90 \\
\bottomrule
\end{tabular}
\end{table}

\subsection{Structural Ablation on Memory Representation and Retrieval}
To study the role of different memory structures and retrieval strategies in \textsc{LatentGraphMem}, we conduct a series of structural ablations summarized in Figure~\ref{fig:graphstructure}. Specifically, we compare our full model with variants that (i) remove subgraph retrieval and reason over the full graph directly, (ii) replace the learned subgraph retriever with a heuristic BFS expansion, and (iii) rely entirely on explicit graph construction and retrieval without latent representations. The results show that directly reasoning over the full graph degrades performance, while heuristic retrieval introduces substantial noise. Fully explicit designs perform better than naive retrieval but remain inferior to our approach, highlighting the necessity of combining latent graph memory with learned, budgeted subgraph retrieval.

\subsection{Inference Time under Long Contexts}
We evaluate the inference-time behavior of different memory systems under increasing context lengths to assess their scalability in long-context settings. Specifically, we measure average inference time on 100 test samples per context length using 1.5B models trained for each method. As shown in Table~\ref{tab:time_vs_context}, A-Mem exhibits highly unstable inference time as the context grows, with latency increasing sharply to 44.65 seconds at 9k tokens. In contrast, \textsc{LatentGraphMem} maintains stable inference time across all evaluated context lengths and closely matches the efficiency of MemGen. This result indicates that, despite exposing explicit subgraph evidence for reasoning, the latent graph memory and fixed retrieval budget in \textsc{LatentGraphMem} prevent inference cost from scaling with the full context length, enabling predictable and scalable inference under long-context workloads.

\begin{figure}[h] 
  \centering
  \includegraphics[width=0.31\textwidth]{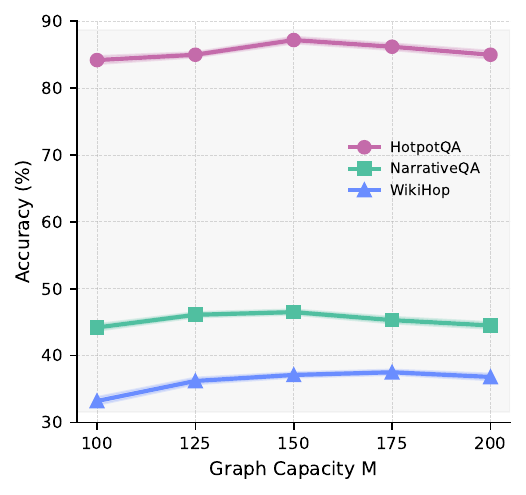}
  \caption{Effect of graph capacity $M$ on answer accuracy across three QA datasets. }
  \label{fig:egc}
\end{figure}

\begin{figure}[!t] 
  \centering
  \includegraphics[width=0.47\textwidth]{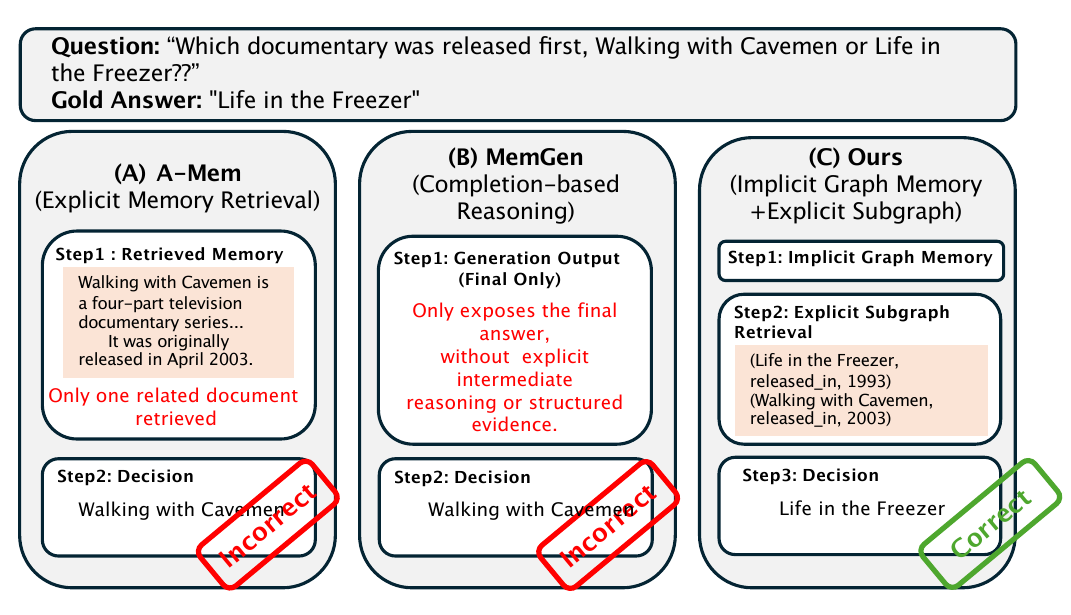}
  \caption{Case Study: Both Explicit and Implicit Construction Baselines Fail, While Our Method Succeeds}
  \label{fig:casestudy}
\end{figure}

\subsection{Effect of Global Graph Capacity}
We analyze the impact of the global graph capacity \(M\) on model performance to understand the trade-off between memory coverage and noise. Figure~\ref{fig:egc} reports results on a 1.5B backbone as we vary \(M\) around the default setting (\(M{=}150\)). We observe that increasing \(M\) consistently benefits WikiHop, while the gains on HotpotQA and NarrativeQA saturate or slightly decline beyond the default value. This behavior aligns with dataset characteristics: WikiHop requires aggregating evidence across a larger set of candidate entities and relations, making broader graph coverage advantageous. In contrast, HotpotQA and NarrativeQA typically involve fewer relevant entities but more localized or narrative-driven evidence, where larger graphs introduce additional noise that can hinder retrieval and reasoning. 

\subsection{Case Study}
We conduct a qualitative case study to examine how different memory paradigms handle sparse and dispersed evidence in long-horizon question answering. The example considers a temporal comparison between two similarly named documentaries, where the relevant release dates are distributed across a long and noisy context.
As illustrated in Figure~\ref{fig:casestudy}, explicit memory retrieval (A-Mem) retrieves evidence for only \emph{Walking with Cavemen} (2003) and fails to retrieve the corresponding date for \emph{Life in the Freezer} (1993), leading to an incorrect comparison. Completion-based latent memory (MemGen) produces a final answer without exposing intermediate evidence, making the underlying temporal reasoning unverifiable and resulting in an incorrect decision. In contrast, \textsc{LatentGraphMem} preserves both entities and their release years in an implicit graph memory and retrieves a compact explicit subgraph containing the relevant dates, enabling a reliable comparison and the correct answer.

\section{Conclusion}
We present \textsc{LatentGraphMem}, a memory framework for long-horizon question answering that integrates latent graph memory with explicit, budgeted subgraph retrieval. By decoupling memory construction from reasoning, our approach combines the stability and efficiency of latent representations with the interpretability and control of explicit evidence exposure. Extensive experiments across multiple long-horizon benchmarks demonstrate that \textsc{LatentGraphMem} achieves stronger and more consistent performance than representative explicit and latent memory baselines while maintaining stable inference time and supporting flexible scaling to larger reasoners. These results highlight the importance of structured yet implicit memory representations for reliable reasoning under long and dispersed contexts.

\section*{Limitations}
While \textsc{LatentGraphMem} is designed for long-horizon reasoning with sparse evidence, its performance depends on the quality of the learned graph construction and retrieval modules. Although we adopt fixed budgets and capacity constraints to ensure stability and efficiency, the optimal configuration may vary across tasks with different evidence structures. In addition, our framework currently focuses on text-based inputs and question answering settings, and extending the approach to other modalities or interactive reasoning scenarios remains an interesting direction for future work.



\bibliography{IMKG}

\appendix

\section{Appendix}
\label{sec:appendix}

\begin{table*}[t]
\centering
\footnotesize
\setlength{\tabcolsep}{4.8pt}
\renewcommand{\arraystretch}{1.12}

\definecolor{tblHead}{gray}{0.92}
\definecolor{tblGroup}{gray}{0.95}
\definecolor{tblTest}{gray}{0.97}
\definecolor{tblTotal}{gray}{0.90}

\newcolumntype{L}[1]{>{\raggedright\arraybackslash}p{#1}}
\newcolumntype{Y}{>{\raggedright\arraybackslash}X}

\newcommand{\longlen}{$^{\uparrow}$}
\newcommand{\shortlen}{$^{\downarrow}$}

\begin{tabularx}{\textwidth}{
  L{0.33\textwidth}  
  L{0.07\textwidth}  
  L{0.08\textwidth}  
  Y                  
  L{0.18\textwidth}  
}
\toprule
\rowcolor{tblHead}
\textbf{Dataset} & \textbf{Split} & \textbf{\#Inst.} & \textbf{Context avg.} & \textbf{Answer avg.} \\
\midrule

\rowcolor{tblGroup}
\multicolumn{5}{l}{\textbf{Training corpora}} \\
TriviaQA \citep{joshi2017triviaqa} & Train & 14{,}679 &
75{,}314 chars (18{,}828 tok)\longlen & 9.9 chars (2.1 tok) \\
QASPER \citep{dasigi2021qasper} & Train & 3{,}598 &
24{,}608 chars (6{,}152 tok) & 6.0 chars (1.0 tok)\shortlen \\
QuALITY \citep{pang2022quality} & Train & 2{,}523 &
27{,}018 chars (6{,}754 tok) & 49.4 chars (12.0 tok)\longlen \\
\addlinespace[2pt]

\rowcolor{tblGroup}
\multicolumn{5}{l}{\textbf{Evaluation corpora}} \\
\rowcolor{tblTest}
HotpotQA \citep{yang2018hotpotqa} & Test & 1{,}000 &
5{,}240 chars (1{,}310 tok) & 11.0 chars (2.7 tok) \\
\rowcolor{tblTest}
NarrativeQA \citep{kocisky2018narrativeqa} & Test & 800 &
3{,}860 chars (965 tok)\shortlen & 26.0 chars (6.5 tok) \\
\rowcolor{tblTest}
WikiHop \citep{welbl2018wikihop} & Test & 800 &
7{,}230 chars (1{,}807 tok) & 12.3 chars (3.1 tok) \\
\midrule

\rowcolor{tblTotal}
Total & -- & 23{,}400 &
5{,}428 chars (1{,}356 tok)$^{\dagger}$ & 16.0 chars (4.0 tok)$^{\dagger}$ \\
\bottomrule
\end{tabularx}

\caption{Datasets used for training and evaluation. ``tok'' denotes token counts under the corresponding base-model tokenizer.
$\uparrow/\downarrow$ mark unusually long/short averages within the table (Context or Answer).
$^{\dagger}$For the total row, we report the average over the test mixture (2,600 instances), consistent with our main evaluation protocol.}
\label{tab:datasets}
\end{table*}

\subsection{Dataset Statistics and Characteristics}
\label{sec:dataset_stats}

This appendix reports detailed statistics and characteristics of the datasets used in our experiments.
Table~\ref{tab:datasets} lists the number of instances and the average lengths of contexts and answers for each dataset.
All statistics are computed on our processed splits: character counts are measured on the preprocessed raw text, and token counts are computed using the corresponding base-model tokenizer to ensure comparability across model configurations.

\paragraph{Training datasets.}
\textbf{TriviaQA}~\citep{joshi2017triviaqa} contains open-domain questions paired with large evidence collections, where contexts are typically long and heterogeneous after preprocessing, making it suitable for training memory mechanisms that must aggregate dispersed evidence.
\textbf{QASPER}~\citep{dasigi2021qasper} is document-grounded QA over full scientific papers, where contexts exhibit structured regularities (e.g., sections and citations) and questions often require integrating information across distant sections.
\textbf{QuALITY}~\citep{pang2022quality} provides multiple-choice questions over long narrative or expository articles, where correct options frequently depend on global comprehension rather than local lexical matches, complementing extractive or short-answer supervision.

\paragraph{Evaluation datasets.}
\textbf{HotpotQA}~\citep{yang2018hotpotqa} and \textbf{WikiHop}~\citep{welbl2018wikihop} are multi-hop benchmarks requiring reasoning across multiple documents or entities, and \textbf{NarrativeQA}~\citep{kocisky2018narrativeqa} focuses on long-form narrative understanding with evidence distributed across stories.
We use fixed test splits from these datasets and do not include them in training.

\paragraph{Test mixture.}
Our aggregate evaluation follows a fixed 2{,}600-instance mixture consisting of 1{,}000 HotpotQA, 800 NarrativeQA, and 800 WikiHop instances.
The \textit{Total} row in Table~\ref{tab:datasets} reports length statistics averaged over this mixture, consistent with our main evaluation protocol.

\subsection{Supplementary Experiment Analysis}
\label{sec:supp_analysis}

\paragraph{Dataset-level trends.}
\textsc{LatentGraphMem} shows the largest accuracy gains on datasets requiring compositional reasoning over dispersed evidence.
It consistently performs best on HotpotQA across all backbones, and its advantage on WikiHop becomes more pronounced as the frozen reasoner scales.
On NarrativeQA, where answers are longer and more abstractive, methods tend to differ more in ROUGE-L than in exact-match accuracy; nevertheless, \textsc{LatentGraphMem} remains competitive in accuracy while still exposing interpretable retrieved subgraphs.

\paragraph{Failure modes of fully explicit graph-based memory.}
Across backbones, fully explicit graph-memory systems (e.g., THEANINE, PREMem, and Mem0/Mem0g) frequently collapse on NarrativeQA, producing near-zero or single-digit accuracy despite occasional competitiveness on HotpotQA.
This pattern suggests that under long horizons, errors introduced by symbolic extraction and retrieval can dominate the downstream reasoning signal, so scaling the frozen reasoner alone does not reliably recover performance.
In contrast, latent-based approaches are more stable under long contexts, and \textsc{LatentGraphMem} further improves robustness by combining latent storage with explicit subgraph exposure.

\end{document}